\documentclass{article}

\usepackage{spconf,amsmath,graphicx}
\usepackage{bm}
\usepackage{url}
\usepackage{hyperref}

\title{Improving Accented Speech Recognition with Multi-Domain Training}

\name{Lucas Maison$^{\star \dagger}$ \qquad Yannick Esteve$^{\star}$}

\address{$^{\star}$ Laboratoire Informatique d'Avignon, Avignon, France \\
$^{\dagger}$ Thales SIX, Multimedia Lab, Gennevilliers, France}

\begin{document}
%\ninept
%
\maketitle
\begin{abstract}
Thanks to the rise of self-supervised learning, automatic speech recognition~(ASR) systems now achieve near human performance on a wide variety of datasets. However, they still lack generalization capability and are not robust to domain shifts like accent variations. In this work, we use speech audio representing four different French accents to create fine-tuning datasets that improve the robustness of pre-trained ASR models. By incorporating various accents in the training set, we obtain both in-domain and out-of-domain improvements. Our numerical experiments show that we can reduce error rates by up to 25\% (relative) on African and Belgian accents compared to single-domain training while keeping a good performance on standard French.
\end{abstract}
\begin{keywords}
automatic speech recognition, multi-domain training, accented speech, self-supervised learning, domain shift
\end{keywords}

\section{Introduction and Related Work}
\label{sec:intro}

In recent years, self-supervised learning~(SSL) has proven useful in a variety of domains, including computer vision, NLP, and speech processing. For the latter, SSL is used to leverage huge amounts of unannotated speech to build powerful speech representations that can be used to tackle downstream tasks like automatic speech recognition~(ASR). Systems pre-trained with an SSL approach and then fine-tuned on a target domain reach state-of-the-art performance. However, there is growing evidence that these systems are not robust to domain shifts~\cite{Hsu2021, gomez2023ATC}. For example, a model fine-tuned on speech uttered with accent A is likely to have reduced performance when evaluated on accent B, although the language is the same in both cases. It is known that adding out-of-domain data to the training dataset improves robustness to domain shifts~\cite{likhomanenko21_interspeech}; however, it is often the case that out-of-domain data is scarce, making it hard to gather amounts sufficient for training. Regarding the peculiar case of accented speech, the vast majority of corpora consist of speech spoken with the dominant accent.

In order to tackle the lack of accented data, we can use data augmentation to artificially increase the amount of training data. There exist several ways to do this, for instance, it is possible to add noise to the training data, modify voice speed, or transform voice by manipulating the vocal-source and vocal-tract characteristics~\cite{Fukuda2018}. Other approaches include applying speaker normalization or anonymization methods in a reverse manner, for example using Vocal Tract Length Perturbation~\cite{Jaitly2013} or voice conversion using X-vectors~\cite{fang2019speaker}.

There exist several methods for reducing the effects of domain shifting. It has been shown that adding new domains to the pre-training dataset improves the model's robustness~\cite{likhomanenko21_interspeech}. In order to avoid expensive retraining of large SSL models, Fan et al. introduce DRAFT~\cite{Fan2022}, a method for learning domain-specific knowledge using residual adapters. Viglino et al. use multi-task learning and accents embeddings~\cite{viglino19_interspeech} to tackle accented speech recognition. In this work, we learn domain-specific knowledge during the fine-tuning stage by carefully designing multi-domain training sets.
\section{Experimental context}
\label{sec:experiments}

\begin{table}
\center
%\resizebox{\columnwidth}{!}{
\begin{tabular}{l|cccc}
\hline
Corpus & Duration & \# Utt. & \# Speakers & Accent \\ \hline
CV~\cite{commonvoice2020}    & 56:11    & 46991         & 2897        & None      \\
AAF~\cite{african_accented_french}    & 13:20    & 11344         & 228         & African\footnotemark  \\
CaFE~\cite{cafe}   & 1:09     & 936           & 12          & Quebec    \\
CFPB~\cite{cfpb}   & 4:07     & 6132          & 9           & Belgian  \\\hline
\end{tabular}%}
\caption{Statistics for the datasets (duration in hours)}

\label{tab:datasets}
\end{table}

\footnotetext{While improper, we re-use the term \textit{African accent} to refer to the mix of accents from Cameroon, Chad, Congo, Gabon, and Niger which compose the dataset.}

\begin{table*}[h!]
\center

\begin{tabular}{l|cccccccccccc|c}
Hours  & 0\%  & 10\% & 20\% & 30\%  & 40\%  & 50\%  & 60\%  & 70\%  & 80\%  & 90\%  & 95\%  & 100\% & FullCV \\ \hline
CV     & 0    & 0.89 & 2.00 & 3.42  & 5.32  & 7.97  & 11.96 & 18.6  & 30.87 & 30.87 & 30.87 & 30.87 & 56.19 \\
AAF    & 7.97 & 7.97 & 7.97 & 7.97  & 7.97  & 7.97  & 7.97  & 7.97  & 7.97  & 3.99  & 1.99  & 0     & 7.97  \\
Total  & 7.97 & 8.86 & 9.97 & 11.39 & 13.29 & 15.94 & 19.93 & 26.57 & 38.84 & 34.86 & 32.86 & 30.87 & 64.16 \\ \hline
\#Utt. & 6780 & 7556 & 8516 & 9831  & 11481 & 13678 & 17068 & 22890 & 33580 & 30186 & 28553 & 26800 & 53771 \\
\#Spk. & 140  & 151  & 166  & 188   & 210   & 248   & 300   & 382   & 574   & 508   & 474   & 434   & 3037 
\end{tabular}
\caption{Statistics for the training sets of our first experiment. Percentages indicate the proportion of CV data in the training set. \textit{FullCV} is a dataset containing AAF together with all the splits of CV.}
\label{tab:training_sets}
\end{table*}

\subsection{Datasets}
\label{subsec:datasets}

We use four different datasets of French speech representing different accents. All the datasets are supplied with transcripts. Their main statistics are reported in Table~\ref{tab:datasets}.

\noindent \textbf{CV} (Common Voice~\cite{commonvoice2020}) is a large crowd-sourced multilingual corpus of read speech. We use the French subset of the CommonVoice~3 database. We use the official splits of the dataset - train: 31h, dev: 12h, test: 13h. This is our reference corpus for (accent-free) French speech.

\noindent \textbf{AAF} (African Accented French~\cite{african_accented_french}) is a corpus of read speech. Speakers originate from five African countries (Cameroon, Chad, Congo, Gabon, and Niger) where French is (one of) the official language(s), however, their accent is clearly audible. We split this dataset as such - train: 8h, dev: 3h, test: 3h. This is our target corpus, that is, we want to obtain the best performance on this dataset.

\noindent \textbf{CaFE} (Canadian French Emotional~\cite{cafe}) is a small corpus of acted emotional speech. Speakers have a distinguishing \textit{Québécois} accent. Due to the low amount of audio of this dataset, we do not split it and use it solely for testing.

\noindent \textbf{CFPB} (\textit{Corpus de Français Parlé à Bruxelles} (Corpus of French as Spoken in Brussels)~\cite{cfpb}) is a small corpus of interviews with Brussels speakers with a Belgian accent. We split this dataset as such - train: 3h, test: 1h.

\subsection{Model}
\label{subsec:model}

We use the following wav2vec~2.0 models from the \textit{LeBenchmark}~\cite{evain2021task} initiative: LB-7K-base and LB-7K-large, which were pre-trained on 7,739 hours of French audio. The \textit{base} variant refers to the standard model architecture from~\cite{baevski2020wav2vec} that has 95 million parameters, while the \textit{large} refers to their larger architecture that presents greater capacity (317 million parameters). We use the \textit{LB-7K} variants of the models since previous work~\cite{maison2022promises} has shown that for this task, models pre-trained using the greater quantity of audio performed best.

Each pre-trained wav2vec~2.0 model acts as a speech encoder, which is fine-tuned for the ASR task together with an additional feed-forward network. This head network consists of three linear layers with 768 or 1,024 neurons for a \textit{base} or \textit{large} model, respectively. Each linear layer is followed by batch normalization and a Leaky ReLU~\cite{Maas2013RectifierNI} activation function. We use dropout with $p=0.15$ between each linear layer.
\noindent At last, a final linear layer projects the output into token space, and log-softmax is applied to obtain probabilities of each token.

\subsection{Training}
\label{subsec:training}

We use the \texttt{SpeechBrain}~\cite{speechbrain} toolkit for all our experiments. All our models are fine-tuned during 50 epochs using the CTC loss. Adam~\cite{adam} and Adadelta~\cite{Zeiler2012ADADELTAAA} optimizers with learning rates $10^{-4}$ and $1.0$ are used to update the weights of the wav2vec 2.0~model and the additional top layers respectively. Learning rates are reduced at each epoch in which the validation loss does not improve.
During training, we apply on-the-fly data augmentation using the \texttt{SpeechBrain} time-domain approximation of the SpecAugment~\cite{Park2019specaugment} algorithm: it disrupts audio speed, and randomly drops chunks of audio and frequency bands.

For fine-tuning we use several different training sets, which are formed using varying amounts of audio data from one or more \textit{speech domains}~(accents). We detail the formation of these training sets in section~\ref{sec:method}. We also use a validation set (dev set) for early stopping; this set is composed of 5 hours of audio, evenly distributed between CV and AAF. We believe that using such a validation set favors the selection of a model with good performance in both domains. Note that we take care of separating by speaker when creating splits of the data; this way validation and testing are always done on unknown speakers.

\subsection{Evaluation}
\label{subsec:evaluation}

We evaluate our trained models on four test sets, which stay identical for all the experiments: CV, AAF, CFPB (test splits), and CaFE (whole set). We use the Word Error Rate~(WER) as our test metric; lower is better. Note that we do not use any language model to avoid introducing a bias during evaluation.
\section{Multi-domain Data Augmentation}
\label{sec:method}

\subsection{Data Augmentation using McAdams coefficient}
\label{subsec:mcadams}

We use data augmentation to increase the amount of accented training data. Our approach consists of the alteration of the McAdams coefficient~\cite{mcadams1984spectral}, and is described in \cite{patino2020speaker} in the context of speaker anonymization. It is based upon simple signal processing techniques and is particularly relevant because it is simple to implement and does not require any additional training data. By applying the McAdams transformation to a speech signal with different McAdams coefficients $\alpha$, we generate new signals with the same transcription but uttered by a pseudo voice with a different timbre.

We use values of $\alpha$ ranging from 0.7 to 1 with a step of 0.1 to generate the augmented dataset; note that using $\alpha=1$ does not change the sample. We do not use values of $\alpha \leq 0.6$ because it deteriorates too much the intelligibility. Applying this augmentation to AAF results in a dataset that is $4\times$ bigger than the original set, leading to a total of $\simeq 32$ hours of speech, matching the audio quantity of CV. We denote this augmented dataset \textbf{AAFaug}.

\subsection{Multi-domain mix of datasets}
\label{subsec:multidom}

We want to study the impact of multi-accent fine-tuning on the recognition of individual accents. To do so we design several experiments which consist of varying the number of domains and their associated quantity of speech data.

In our first experiment, we create train sets using various amounts of CV and AAF. Starting with a set containing only AAF, we keep increasing its size by including increasingly larger subsets of CV. In a reverse manner, we also create train sets by starting with CV and progressively adding AAF. We denote these sets \textbf{CV - \bm{$x$}\%}, where $x$ represents the proportion of CV data in the set. In addition, we create a \textit{FullCV} train set consisting of the whole CV dataset (train, dev, test) together with the AAF train split. See Table~\ref{tab:training_sets} for details on the individual training sets.

In a second experiment, we select the best train sets from experiment 1 (i.e. the sets which led to the lowest WER) and combine them with either CaFE (whole) or CFPB (train split). This leads to slightly larger train sets which have three accents instead of two.

In our third and last experiment, we repeat experiment 1 using a fixed number of hours in the training set. To do so, we make use of the augmented dataset AAFaug~\ref{subsec:mcadams}. Similarly to experiment 1, we define 11 train sets, starting with a 31h subset of AAFaug, and gradually replacing more and more AAFaug data by CV data with a 10\% proportion increment, ending with the full CV train set. Finally, we create two larger training sets using the augmented data, namely CV $\cup$ AAFaug and FullCV $\cup$ AAFaug.
\section{Results and Discussion}
\label{sec:results}

\begin{figure}[h]
    \begin{center}
    \includegraphics[width=\columnwidth]{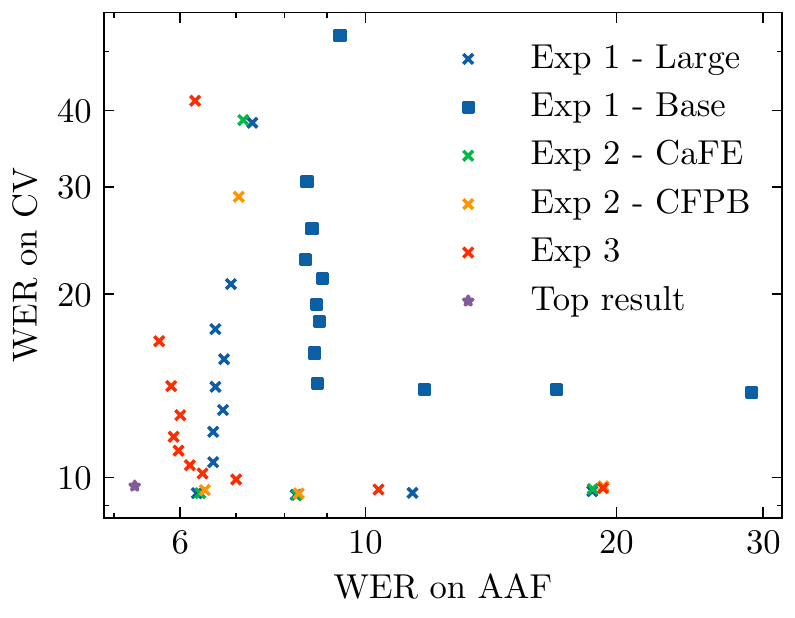}
    \caption{A representation of the performance trade-off between CV and AAF. Each point represents a model fine-tuned on a different train set. All models are \textit{Large} unless specified otherwise. Details on the experiments can be found in section~\ref{subsec:multidom}.}
    \label{fig:wer_cv_aaf_tradeoff}    
    \end{center}
\end{figure}

In this section, we summarize and analyze the results of our experiments. Numerical results are reported in Table~\ref{tab:results}; for the sake of brevity, we only list the best training sets. All models use the \textit{Large} architecture unless specified otherwise.

\subsection{Experiment 1: Mixing two domains}

We study the impact of mixing two domains during training on performance. We observe U-shaped curves of WER with respect to CV proportion in the training set, meaning that WER first decreases when adding CV to the mix (from 0 to 80\%), then slightly increases again when removing AAF (from 80 to 100\%). Going from one to two domains causes both in-domain and out-of-domain improvements. 
When adding a new domain B to a train set containing only domain A, not only do we improve on domain B as expected, but we also improve on domain A and on unseen domains C and D as well. We obtain our best results on CV~(9.4 WER) and on CaFE~(36.8 WER) using CV 90\% and CV 50\% respectively, while also reaching competitive performance on the other test sets compared to the baselines. Remarkably, training on a CV/AAF mixed set allows us to reach WER of 34.6 and 36.8 on CFPB and CaFE respectively, which is a lot better than the scores reached when training on CV or AAF alone, see table~\ref{tab:results}.

\noindent We observe similar trends on base models although with significantly higher WER (+40\% on average relative to large models). However, we see lesser improvements from the domain augmentation compared to large models. We believe the base model is less able to take advantage of multi-domain due to its reduced capacity.

It should be noted that the \emph{LB-7K} models were pre-trained using several different corpora, including AAF and CaFE~(see~\cite{evain2021task}). However, this does not prevent the catastrophic forgetting phenomenon which happens during fine-tuning. For instance, the LB-7K-large model fine-tuned on CV scores 18.7 on AAF, despite AAF being included in the pre-training data. This highlights the importance of mixing domains during fine-tuning.

\subsection{Experiment 2: Mixing three domains}

We select the five best train sets from experiment 1, which are CV 0\%~(i.e. AAF only), CV 80\% (i.e. CV $\cup$ AAF), CV 90\%, CV 100\% (i.e. CV only) and \textit{FullCV}. When adding CaFE to those sets, we observe no significant changes on CV or AAF, as we can see in figure~\ref{fig:wer_cv_aaf_tradeoff}. We observe the same trend with CFPB; however, its addition results in a massive boost on the CFPB test set~(-30\% WER). The best performance on CFPB is reached on the train set CV 90\% $\cup$ CFPB, with a WER of 24.8 and very good scores on CV and AAF.

\noindent Moreover, the best average WER on all four test sets is reached on the train set CV 80\% $\cup$ CFPB, see Table~\ref{tab:results}, last row. This shows the benefits of training on multiple domains, as we can leverage speech from other domains to improve on each domain with respect to the single-domain baseline.

\begin{figure}[h]
    \begin{center}
    \includegraphics[width=\columnwidth]{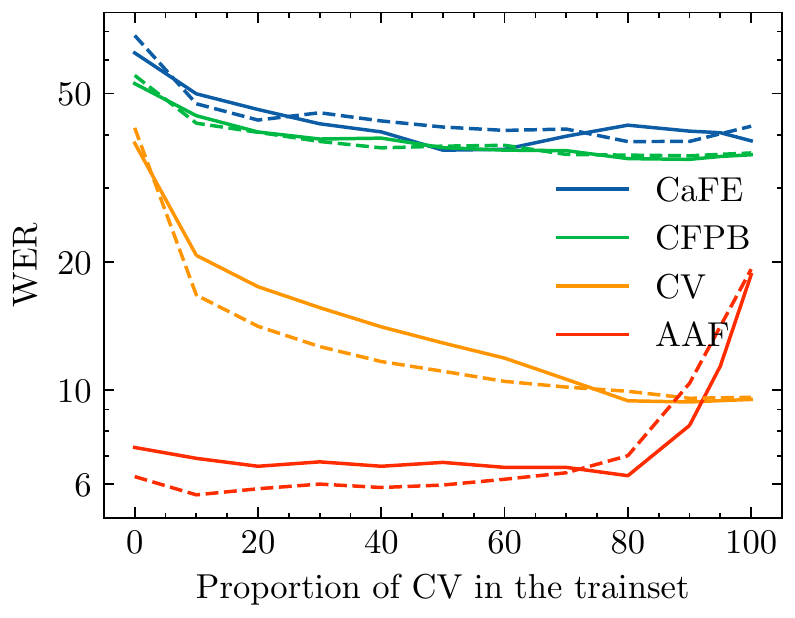}
    \caption{Graph of the WER with respect to the CV proportion. Plain and dashed lines correspond to experiments 1 and 3 respectively (without or with augmentation).}
    \label{fig:results_augmentation}    
    \end{center}
\end{figure}

\subsection{Experiment 3: using data augmentation}

We repeat experiment 1 using an augmented AAF dataset. Results are shown in figure~\ref{fig:results_augmentation}. We clearly see the advantage of using the augmented train set: performance improves on CV and AAF while staying nearly identical on CaFE and CFPB. This improvement is also visible in figure~\ref{fig:wer_cv_aaf_tradeoff}, bottom-left corner. In addition, this figure highlights the existing trade-off between error rates on different domains. When training using a fixed number of hours, an improvement on one domain usually comes at the expense of a (possibly slight) degradation in another domain.

\noindent The best performance on AAF among these 31h sets is reached using CV 10\% (which comprises 28h of African speech) with a WER of 5.7, but its performance is poor on CV~(16.7 WER). However, if we train on CV $\cup$ AAFaug (63h of speech), we obtain a WER of 5.3 and 9.7 on AAF and CV respectively (see "top result" in figure~\ref{fig:results_augmentation}), which is our best score on African accent.

These results on augmentation are encouraging: thanks to it, we are able to significantly reduce the WER on African accent ($-14$\% WER, relative) compared to the best result without augmentation. However, the augmentation method we describe (which is based on the variation of the McAdams coefficient) may not be particularly relevant to the task. Indeed, we ran an ablation study, training using only on-the-fly SpecAugment on the same quantity of data, and obtained nearly the same results. Further investigation is needed to assess more precisely the impact of the different augmentation methods on ASR performance, and the relevance of McAdams pseudo-voices.

Finally, we discuss the impact of using a substantially larger CV domain for training, forming the \textit{FullCV} family of train sets, which are the largest ones. Models trained on these sets tend to perform better on average on CaFE~($-6.1$\%) and CFPB~($-1.4$\%) compared to those trained on the second biggest sets, but also show a performance degradation on AAF~($+1.7$\%). Besides, no model trained on \textit{FullCV} sets has reached top performance on any of the test sets. This is likely due to the discrepancies between the accents in the training data, and indicates that blindly maximizing the amount of training data may not always be the best choice for minimizing WER, neither across domains nor for a single domain.

\begin{table}
\center
\begin{tabular}{l|cccc}
Train set         & CV            & AAF           & CFPB           & CaFE           \\ \hline
CV                & 9.5           & 18.71         & 35.86          & 38.68          \\
AAF               & 38.2          & 7.32          & 52.78          & 62.34          \\
CFPB              & 45.1          & 44.28         & 32.59          & 67.41          \\ \hline
CV 90\%           & \textbf{9.37} & 8.25          & 34.98          & 40.77          \\
CV 100\% $\cup$ AAFaug & 9.69          & \textbf{5.29} & 37.2           & 44.69          \\
CV 90\% $\cup$ CFPB    & 9.42          & 8.32          & \textbf{24.79} & 41.77          \\
CV 50\%           & 12.91         & 6.75          & 37.27          & \textbf{36.76} \\ \hline
CV 80\% $\cup$ CFPB    & 9.54          & 6.42          & 25.21          & 41.09         
\end{tabular}
\caption{Results on the different test sets. Each line corresponds to a different training set. Rows contain, in that order: single-domain train sets; best train sets for each test set; train set with the best average score. The best results in each column are shown in bold}

\label{tab:results}
\end{table}
\section{Conclusion}
\label{sec:conclusion}

Recognition of accented speech remains a challenging task. In this work, we showed the positive influence of using multi-domain fine-tuning datasets both on in-domain and out-of-domain accents. We experimented with various ways of combining training sets and achieved remarkable gains on our evaluation accents compared to single-domain baselines. We leave as future work the collection and evaluation of new accents, as well as the experimentation of new methods for generating accented speech for low-resource accents.

% References should be produced using the bibtex program from suitable
% BiBTeX files (here: strings, refs, manuals). The IEEEbib.bst bibliography
% style file from IEEE produces unsorted bibliography list.
% -------------------------------------------------------------------------
\bibliographystyle{IEEEbib}
\bibliography{refs}

\end{document}